\documentclass{article}

\usepackage[preprint]{neurips_2026}

\usepackage{times}
\usepackage{amsmath,amssymb,amsthm}
\usepackage{graphicx}
\usepackage{booktabs}
\usepackage{hyperref}
\usepackage{cleveref}
\usepackage{xspace}
\usepackage{enumitem}
\usepackage{array}
\usepackage{makecell}

\crefname{assumption}{Assumption}{Assumptions}

\usepackage{pifont}

\title{Hamiltonian World Models: Rethinking World Models from a Physical Dynamics Perspective}

\title{Physically Native World Models: A Hamiltonian Perspective on Generative World Modeling}

\author{
  Sen Cui\thanks{Corresponding author. Email: \texttt{cuis@mail.tsinghua.edu.cn}} \\
  Tsinghua University \\
  \And
  Jingheng Ma \\
  Tsinghua University
}

\begin{document}

\maketitle

\begin{abstract}
World models have recently re-emerged as a central paradigm for embodied intelligence, robotics, autonomous driving, and model-based reinforcement learning. However, current world model research is often dominated by three partially separated routes: 2D video-generative models that emphasize visual future synthesis, 3D scene-centric models that emphasize spatial reconstruction, and JEPA-like latent models that emphasize abstract predictive representations. While each route has made important progress, they still struggle to provide physically reliable, action-controllable, and long-horizon stable predictions for embodied decision making. In this paper, we argue that the bottleneck of world models is no longer only whether they can generate realistic futures, but whether those futures are physically meaningful and useful for action. We propose \emph{Hamiltonian World Models} as a physically grounded perspective on world modeling. The key idea is to encode observations into a structured latent phase space, evolve the latent state through Hamiltonian-inspired dynamics with control, dissipation, and residual terms, decode the predicted trajectory into future observations, and use the resulting rollouts for planning. We discuss how Hamiltonian structure may improve interpretability, data efficiency, and long-horizon stability, while also noting practical challenges in real-world robotic scenes involving friction, contact, non-conservative forces, and deformable objects.
\end{abstract}

\section{Introduction}
\label{sec:intro}

The idea of learning internal simulators has regained prominence as a unifying objective across embodied AI, robotics, and decision
making. This renewed interest is driven by two converging trends. First, large-scale foundation models have demonstrated strong semantic competence in language, vision, and multimodal reasoning \citep{radford2021clip,alayrac2022flamingo,openai2023gpt4,driess2023palme,brohan2023rt2}. Second, embodied agents require predictive models that go beyond recognizing the current scene: they must anticipate how the world will change under their own actions. A robot, for example, should not only identify an object, but also predict whether pushing, grasping, or releasing it will lead to stable manipulation, collision, or task failure. This makes world modeling a natural bridge between perception, prediction, and decision making.

The classical motivation for world models is to provide an internal simulator of the environment, enabling agents to imagine possible futures before acting. Early work showed that learned latent models can support control by compressing observations into compact predictive representations \citep{ha2018worldmodels}. Model-based reinforcement learning further demonstrated that agents can plan or optimize policies inside learned latent dynamics, reducing the need for costly real environment interaction \citep{hafner2019planet,hafner2020dreamer,hafner2023dreamerv3}. In robotics, visual foresight and video prediction have been used to evaluate candidate actions before execution \citep{finn2017deepvisualforesight,ebert2018visualmpc}. In autonomous driving and interactive simulation, generative world models such as GAIA-1 and Genie suggest that future scene generation can be conditioned on actions, text, or latent controls \citep{hu2023gaia,bruce2024genie}. Together, these developments have made ``generating the future'' an increasingly important research direction.

However, generating future video is not equivalent to understanding the physical world. Modern video generative models can synthesize visually plausible sequences, but visual plausibility alone does not guarantee physical validity. A generated trajectory may look realistic over a short horizon while gradually violating object permanence, contact consistency, momentum conservation, or action-effect causality. This distinction is especially important for robotics and autonomous systems, where an incorrect future prediction can directly mislead planning and execution. For entertainment-oriented video synthesis, perceptual realism may be sufficient; for embodied decision making, the predicted future must also be dynamically feasible, controllable, and reliable.

A key limitation of many current video world models is that their temporal evolution is learned through highly expressive but weakly constrained neural sequence models. In a typical latent formulation, the future state is predicted as
\begin{equation}
    z_{t+1} = f_\theta(z_t, a_t),
    \label{eq:1}
\end{equation}
where $z_t$ is a learned latent state and $a_t$ is an action. Although this formulation is flexible and compatible with recurrent latent dynamics, diffusion-based video predictors, Transformers, and structured state-space models 
\citep{ha2018worldmodels,hafner2019planet,ho2022videodiffusion,blattmann2023videoldm,vaswani2017attention,gu2022s4,gu2023mamba}, 
it does not specify what physical quantities the latent state represents or what structure the transition should preserve. As a result, the model may learn short-term correlations in pixel or latent space without acquiring stable physical dynamics. This often leads to compounding errors, long-horizon drift, poor out-of-distribution generalization, and high data requirements.

This limitation suggests that the bottleneck of world models is no longer only visual fidelity. The more fundamental question is whether the generated future is physically coherent and useful for action. A world model should not merely predict what the next frames look like; it should represent how objects, agents, and their interactions evolve under interventions. Prior work on model-based control and robotic video prediction has already shown the practical value of imagined rollouts for planning \citep{finn2017deepvisualforesight,ebert2018visualmpc,hafner2020dreamer,hafner2023dreamerv3}. Yet for such rollouts to support reliable decision making, the latent dynamics must be stable, interpretable, and constrained by the regularities of the physical world.

We therefore argue that the next stage of world model research should move from visual generation toward physical structure modeling. In this view, video generation is an observation-level interface, while the core of a world model is its latent dynamics. The internal state should encode entities, relations, controllable variables, and physically meaningful structure; the transition should reflect constraints such as geometry, energy, contact, and action-conditioned causality; and the generated future should be treated as a rendering of an underlying state evolution. This perspective is consistent with broader efforts in structured dynamics learning, object-centric representation learning, and physics-informed machine learning \citep{battaglia2016interaction,greydanus2019hamiltonian,toth2020hamiltonian,raissi2019physics,sanchezgonzalez2020learning,kipf2020contrastive}.

Motivated by this perspective, we propose to rethink world models through the lens of Hamiltonian dynamics. Hamiltonian mechanics provides a principled framework for describing physical evolution in phase space through generalized coordinates, momenta, and an energy function. Rather than asking a video model to discover physical laws purely from data, a Hamiltonian world model leverages energy-based dynamics in a learned phase space to govern latent evolution. This offers a path toward world models that are not only generative, but also physically interpretable, data-efficient, and stable under long-horizon rollout.

\section{Existing Three Routes Toward World Models}
\label{sec:related_work}

World model research has recently developed along several partially overlapping but conceptually distinct routes. In this paper, we group current approaches into three representative paradigms: \emph{2D video-generative world models}, \emph{3D scene-centric world models}, and \emph{JEPA-like latent world models}. This classification is not meant to be exhaustive, but it clarifies what different routes primarily optimize for. The 2D video route emphasizes future observation synthesis, the 3D route emphasizes spatial and geometric reconstruction, and the JEPA-like route emphasizes predictive abstraction in latent representation space.

\paragraph{2D video-generative world models.}
The first route treats world modeling as conditional future video generation. Given past observations, actions, language, maps, or other conditioning signals, the model predicts future visual observations: $p_\theta(o_{t+1:t+H} \mid o_{\leq t}, a_{t:t+H-1}, c)$, where $o_t$ denotes an observation, $a_t$ denotes an action or control input, and $c$ denotes additional context. This route is closely related to recent advances in video diffusion and latent video generation, which have significantly improved the quality and temporal coherence of generated videos \citep{ho2022videodiffusion,blattmann2023videoldm}. In world modeling, GAIA-1 formulates autonomous driving world modeling as controllable generative sequence modeling over video, text, and action inputs \citep{hu2023gaia}, while Genie demonstrates that action-controllable interactive environments can be learned from large-scale videos \citep{bruce2024genie}. The strength of this route is that it provides an intuitive observation-level interface: the future is represented in the same visual space in which agents perceive the world. Its limitation is that a video may look realistic while still violating contact, momentum, action-effect consistency, or long-horizon stability.

\paragraph{3D scene-centric world models.}
The second route emphasizes explicit spatial structure. Instead of representing the world primarily as a sequence of 2D frames, 3D scene-centric models aim to recover geometry-aware representations that support novel view synthesis, reconstruction, and spatial reasoning. Neural Radiance Fields model scenes as continuous volumetric radiance fields and have become a foundation for view-consistent scene representation \citep{mildenhall2020nerf}. 3D Gaussian Splatting further improves real-time rendering efficiency and visual quality using explicit 3D Gaussian primitives \citep{kerbl20233dgs}. This route is important because embodied agents must operate in three-dimensional space, and purely 2D video prediction may fail to capture view consistency, occlusion structure, and metric geometry. However, 3D reconstruction alone is not yet a complete world model. Many 3D scene representations are better suited to static or quasi-static reconstruction than to action-conditioned physical evolution. They can answer where objects are and how a scene looks from another view, but they do not necessarily answer how objects move, collide, deform, or respond to robot actions.

\paragraph{JEPA-like latent world models.}
The third route focuses on predictive representation learning in latent space. Instead of reconstructing pixels, Joint-Embedding Predictive Architectures learn to predict abstract representations of missing or future inputs \citep{assran2023ijepa,bardes2024vjepa}. This design is motivated by the idea that intelligent systems should predict high-level features rather than every low-level pixel. Compared with pixel-level generation, latent prediction can be more compact, more scalable, and less dominated by irrelevant visual details. This makes JEPA-like models attractive as a route toward abstract world representations. However, because these models are primarily designed for representation learning rather than explicit physical simulation, their latent variables are not necessarily aligned with physical quantities such as position, velocity, momentum, energy, or contact constraints. Therefore, while JEPA-like models may provide strong predictive abstractions, additional structure is needed if such representations are to support physically grounded planning and control.


\section{Limitations of Current World Models}
\label{sec:limit}

Despite rapid progress, current world models still face several fundamental limitations when used for embodied decision making. These limitations do not simply reflect insufficient model scale or visual quality; rather, they may reveal a deeper mismatch between generating plausible observations and modeling physically reliable world dynamics.

\paragraph{Visual fidelity does not guarantee physical validity.}
As noted above, video world models optimize perceptual realism more directly than physical consistency. This becomes critical in robotics and autonomous driving, where a generated future is useful only if it corresponds to a dynamically feasible outcome.

\paragraph{Data efficiency remains limited when physical priors are absent.}
A purely data-driven world model must infer physical regularities from observed trajectories. This can require large and diverse datasets covering different objects, masses, contacts, materials, actions, and environmental conditions. In robotics, such data are expensive to collect, and unsafe exploration may be impractical. Model-based learning improves sample efficiency by reusing imagined trajectories \citep{ha2018worldmodels,hafner2020dreamer,hafner2023dreamerv3}, but the quality of this advantage depends on the reliability of the learned model.

\paragraph{Latent transitions are often weakly structured.}
Many latent world models represent dynamics with a generic neural transition that is learned end-to-end. This has been successful in model-based reinforcement learning \citep{hafner2019planet,hafner2020dreamer,hafner2023dreamerv3}, but it does not specify what physical quantities the latent variables encode or what structure the transition should preserve. As a result, the latent state may entangle appearance, semantics, motion, and task-specific features without representing coordinates, velocities, momenta, forces, or constraints. Many modern video world models still rely primarily on high-capacity sequence modeling rather than explicit physical structure.

\paragraph{Long-horizon rollout remains vulnerable to compounding errors.}
World models are most valuable when they can be rolled out over multiple steps for planning. Yet learned dynamics models are well known to suffer from compounding prediction errors: small one-step inaccuracies may accumulate and produce unreliable long-horizon trajectories \citep{talvitie2017self,janner2019mbpo}. This issue is especially severe when the model is queried on states or actions outside the training distribution during planning.

\paragraph{Generated futures are not always aligned with action usefulness.}
A world model for embodied intelligence should support action selection, not merely future synthesis. Visual foresight methods explicitly use action-conditioned video prediction for model predictive control \citep{finn2017deepvisualforesight,ebert2018visualmpc}, while latent imagination methods train policies from predicted trajectories \citep{hafner2020dreamer,hafner2023dreamerv3}. These works show that imagined futures can be useful for control, but they also reveal a key requirement: the prediction must preserve the causal effect of actions. If a model generates visually coherent futures that are weakly conditioned on actions, or if it fails to distinguish feasible from infeasible action outcomes, it becomes unreliable for planning. This limitation is particularly important for contact-rich manipulation, where small differences in force, momentum, or contact geometry can determine task success.

\section{Redefining World Models for Embodied Intelligence}
\label{sec:define}


The limitations of current world models suggest that the concept itself needs to be sharpened. We hypothesize a world model as a structured, action-conditioned, data-efficient, and physically grounded generative dynamical system that represents the current world state, predicts how it evolves under interventions, and supports reliable decision making. 
This definition is broader than classical latent dynamics models \citep{ha2018worldmodels,hafner2019planet,hafner2020dreamer,hafner2023dreamerv3}, but also more constrained than generic video generation models \citep{ho2022videodiffusion,blattmann2023videoldm,hu2023gaia,bruce2024genie}. We do not claim these properties are universally necessary for all world modeling tasks; rather, we argue that their combination provides a useful design target for physically situated agents, and that Hamiltonian dynamics offers one principled way to satisfy them.

\paragraph{Physical grounding.}
For physical agents, not every visually plausible future is dynamically feasible. A world model should respect physical regularities such as geometry, contact, inertia, energy variation, and interaction constraints. This does not mean that the model must be a hand-coded simulator. Rather, it should incorporate physical structure as an inductive bias in its representation, transition, or objective. For world models, physical grounding is especially important because imagined trajectories may be directly used for planning and control.

\paragraph{Data-efficient through physical priors.}
Purely data-driven video or latent prediction can therefore become inefficient when the model must infer basic physical structure only from observed trajectories. 
A more scalable world model should incorporate physical priors that constrain the hypothesis space of possible dynamics. 
Physical priors may take the form of structured representations, relational inductive biases, geometric constraints, energy-based dynamics, or physics-informed
objectives.\citep{battaglia2016interaction,sanchezgonzalez2020learning,raissi2019physics,greydanus2019hamiltonian,cranmer2020lagrangian}. For embodied world modeling, data efficiency should therefore be treated as a core design requirement: the model should learn from data, but it should not require data to rediscover every basic law of physical interaction.

\paragraph{Compositional structure.}
The physical world is compositional: objects, agents, tools, surfaces, and constraints interact through relations. A world model should therefore avoid representing the entire scene as an unstructured global vector whenever possible. Instead, it should encode entities and their interactions through structured latent variables,
\begin{equation}
    z_t = \{z_t^i\}_{i=1}^{N},
\end{equation}
where each latent component may correspond to an object, an agent, or a dynamically relevant part of the scene. This view is supported by object-centric representation learning, interaction networks, and graph-based physical simulators, which show that relational structure improves generalization in physical domains \citep{battaglia2016interaction,sanchezgonzalez2020learning,kipf2020contrastive}. A structured world model is easier to inspect and compose, and more suitable for modeling physical interactions than a monolithic latent representation.

\paragraph{Temporal consistency under rollout.}
A static representation of the scene is insufficient for embodied agents. The defining property of a world model is its ability to roll the world forward:
\begin{equation}
    z_{t+1} = T_\theta(z_t, a_t),
\end{equation}
where $T_\theta$ denotes a learned transition and $a_t$ denotes an action or intervention. This transition-centric view is central to model-based reinforcement learning and planning \citep{sutton1991dyna,hafner2019planet,hafner2020dreamer,schrittwieser2020muzero,hafner2023dreamerv3}. However, for embodied intelligence, evolvability should mean more than one-step prediction. A useful world model must remain stable under multi-step rollout, generalize beyond the training distribution, and preserve the relevant structure of the environment over time.

\paragraph{Causal action conditioning.}
Embodied agents do not passively observe the world; they intervene in it. Therefore, a world model must answer counterfactual questions of the form: what will happen if the agent executes action $a_t$? Formally, the model should estimate
\begin{equation}
    p_\theta(o_{t+1:t+H} \mid o_{\leq t}, a_{t:t+H-1}),
\end{equation}
or, at the latent level,
\begin{equation}
    p_\theta(z_{t+1:t+H} \mid z_t, a_{t:t+H-1}).
\end{equation}
This requirement distinguishes embodied world modeling from passive video prediction. Robotic visual foresight and latent imagination methods have demonstrated that action-conditioned prediction can support planning and policy learning \citep{finn2017deepvisualforesight,ebert2018visualmpc,hafner2020dreamer,hafner2023dreamerv3}. Without strong action conditioning, a model may generate plausible futures but fail to represent the causal consequences of the agent's choices.


We therefore summarize these requirements compactly:
\begin{align}
    z_t &= E_\theta(o_{\leq t}), \\
    \hat{z}_{t+1:t+H} &= T_\theta(z_t, a_{t:t+H-1}), \\
    \hat{o}_{t+1:t+H} &\sim D_\theta(\hat{z}_{t+1:t+H}), \\
    a_{t:t+H-1}^\ast &= \arg\max_{a_{t:t+H-1}} U(\hat{z}_{t+1:t+H}, \hat{o}_{t+1:t+H}),
\end{align}
where $E_\theta$ encodes observations, $T_\theta$ evolves latent dynamics, $D_\theta$ renders future observations, and $U$ evaluates decision utility. This formulation separates four roles that are often conflated: perception, dynamics, generation, and planning.
\vspace{-.2cm}
\section{Physical Backbone: Hamiltonian Dynamics for World Models}
\label{sec:hwm}

If the core of a world model is its latent dynamics, then the choice of dynamical structure becomes central. For embodied intelligence, a useful world model should not only predict future observations, but also represent how physical states evolve under constraints, interactions, and interventions. Hamiltonian mechanics offers a natural candidate for such a backbone because it describes physical evolution through phase-space variables, energy functions, and structure-preserving dynamics. This makes it particularly suitable for rethinking world models as physically grounded generative dynamical systems rather than unconstrained video predictors.

\paragraph{Phase space provides a natural latent state.}
Many video world models rely on latent variables whose physical meaning is unclear. Hamiltonian mechanics instead begins with a structured state representation:
\begin{equation}
    z_t = [q_t, p_t],
\end{equation}
where $q_t$ denotes generalized coordinates and $p_t$ denotes generalized momenta. This phase-space view is well aligned with the needs of embodied world modeling: an agent must reason not only about where objects are, but also about how they are moving and how they may respond to future actions. Prior work on Hamiltonian Neural Networks and Hamiltonian Generative Networks shows that phase-space structure can provide useful inductive bias for learning dynamics, including from high-dimensional observations \citep{greydanus2019hamiltonian,toth2020hamiltonian}. Compared with an unstructured latent vector, a phase-space latent state makes the transition model more interpretable and more directly connected to physical evolution.

\paragraph{Energy functions induce structured dynamics.}
A key strength of Hamiltonian modeling is that the dynamics are derived from a scalar energy function rather than predicted by an arbitrary vector field. Given a Hamiltonian $H(q,p)$, the evolution follows
\begin{equation}
    \dot{q} = \frac{\partial H}{\partial p},
    \qquad
    \dot{p} = -\frac{\partial H}{\partial q}.
\end{equation}
Equivalently, with $z=[q,p]$,
\begin{equation}
    \dot{z}=J\nabla_z H(z),
    \qquad
    J=
    \begin{bmatrix}
        0 & I \\
        -I & 0
    \end{bmatrix}.
\end{equation}
This formulation changes the learning problem: instead of directly fitting $z_{t+1}=f_\theta(z_t,a_t)$, the model learns an energy landscape that generates dynamics through its gradients. HNNs demonstrate that this design can improve generalization over unconstrained neural dynamics in physical systems \citep{greydanus2019hamiltonian}. For world models, this suggests a principled way to constrain latent transitions so that future rollouts are guided by physical structure rather than short-term statistical correlation alone.

\paragraph{Hamiltonian structure supports long-horizon stability.}
World models are useful only if their rollouts remain reliable over multiple steps. However, generic learned dynamics often suffer from compounding errors and long-horizon drift \citep{talvitie2017self,janner2019mbpo}. Hamiltonian systems provide a different bias: their dynamics preserve phase-space structure, and when combined with symplectic integration, they can maintain more stable qualitative behavior over long rollouts. Structure-preserving numerical methods have long been used for physical simulation 
\citep{hairer2006geometric}. 
Recent neural approaches, including Hamiltonian Neural Networks 
\citep{greydanus2019hamiltonian}, Hamiltonian Generative Networks 
\citep{toth2020hamiltonian}, and Symplectic ODE-Net 
\citep{zhong2020symplectic}, further show that related structure-preserving principles can be incorporated into learned dynamics. This is especially relevant for embodied planning, where small errors in early imagined states may otherwise lead to unreliable action choices.

\paragraph{Physical priors improve data efficiency.}
As argued in Section~\ref{sec:define}, physical priors are essential for data efficiency\citep{raissi2019physics,battaglia2016interaction,sanchezgonzalez2020learning,greydanus2019hamiltonian,cranmer2020lagrangian}. Hamiltonian dynamics specifically reduce the learning burden by replacing unconstrained local transitions with a global energy landscape; rather than fitting trajectories independently, the model learns a single structure that generalizes across initial conditions.

\paragraph{Hamiltonian dynamics makes latent evolution more interpretable.}
Interpretability is important for world models because their predictions may be used for planning, safety checking, or failure diagnosis. In an unconstrained latent transition, it is often unclear whether prediction errors arise from perception, dynamics, control conditioning, or rendering. A Hamiltonian formulation separates these roles more clearly. The latent variables can be interpreted as phase-space coordinates, the Hamiltonian can be interpreted as an energy-like scalar, and the rollout can be inspected through phase trajectories, energy variation, and interaction terms. Hamiltonian Generative Networks demonstrate that such structure can be learned from pixels, suggesting that physically interpretable latent dynamics need not require direct access to low-dimensional ground-truth states \citep{toth2020hamiltonian}. For embodied world models, this interpretability is valuable because it enables diagnosis of whether an imagined future is physically plausible or merely visually plausible.

\paragraph{Hamiltonian structure can be extended beyond ideal conservative systems.}
A common concern is that real-world robotic scenes are not closed conservative systems. They involve actuation, friction, damping, impacts, contact switches, and unmodeled visual factors. This concern is valid: a useful Hamiltonian world model should not assume perfect energy conservation. Instead, Hamiltonian dynamics should be treated as a structural backbone that can be extended with control and dissipation. Prior work on controlled Hamiltonian dynamics and Symplectic ODE-Net supports the feasibility of incorporating control inputs into Hamiltonian-inspired neural dynamics \citep{zhong2020symplectic}. This extension is crucial: the goal is not to force the real world into an ideal conservative system, but to use Hamiltonian structure as a physically meaningful inductive bias.
\section{An Architectural View of Hamiltonian World Models}
\label{sec:hwm_2}
\begin{figure*}[t]
    \centering
    \includegraphics[ width=\textwidth]{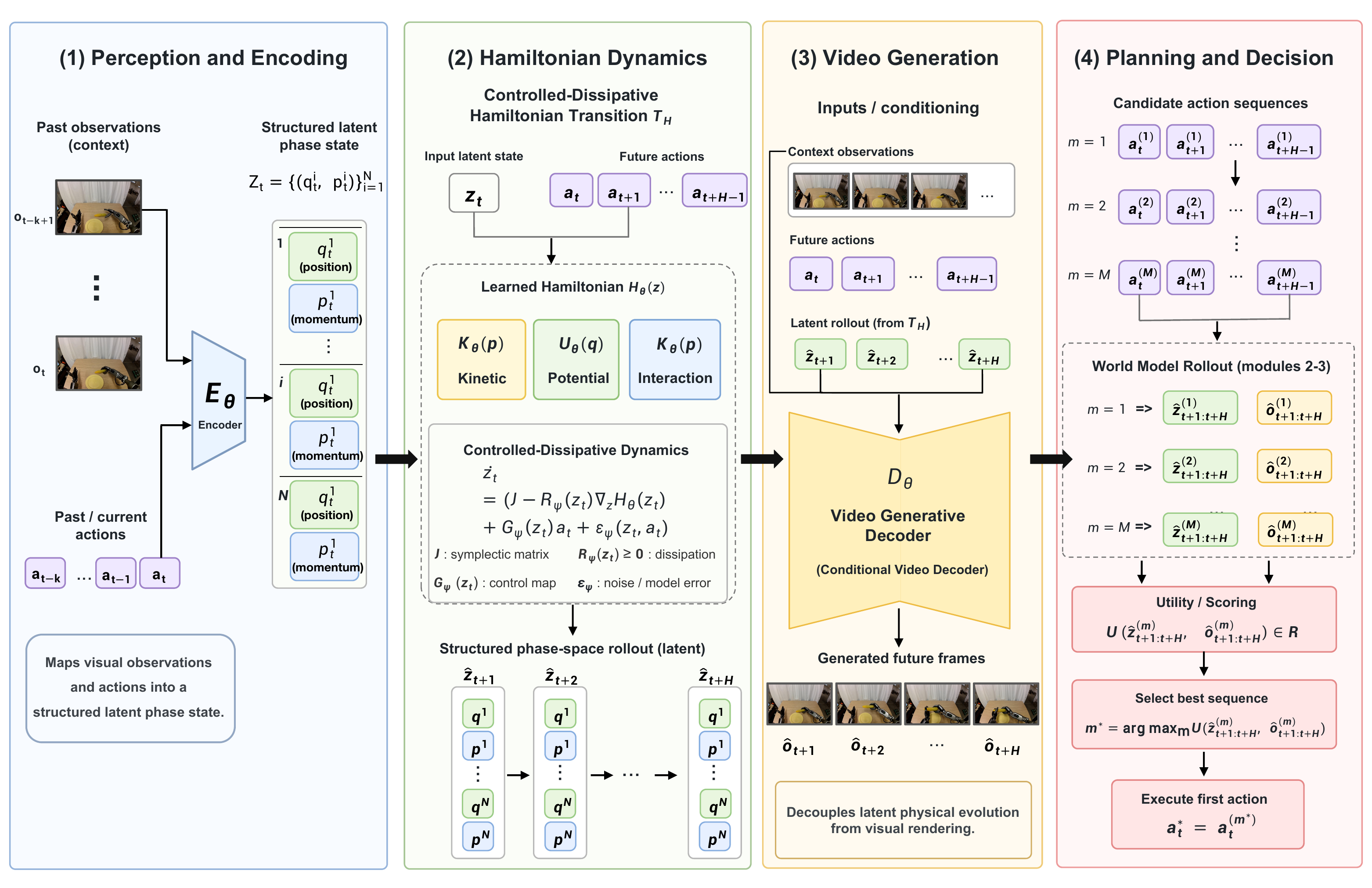}
    \setlength{\abovecaptionskip}{-15pt}
    \setlength{\belowcaptionskip}{-15pt} 
    \caption{
    Overview of the proposed Hamiltonian World Model (HWM) architecture.
    \textbf{(1) Perception and Encoding:} past observations and actions are encoded into a structured latent phase state;
    \textbf{(2) Hamiltonian Dynamics:} the latent state is evolved through Hamiltonian-inspired dynamics;
    \textbf{(3) Video Generation:} the predicted latent trajectory is decoded into future visual observation;
    \textbf{(4) Planning and Decision:} candidate action sequences are rolled out through the world model.
    }
    \label{fig:hwm_architecture}
\end{figure*}
As we note in Section~\ref{sec:hwm} , for embodied intelligence, the world model must connect perception, dynamics, generation, and decision making. We therefore outline a possible architecture for \emph{Hamiltonian World Models}: a physically grounded generative dynamical system that encodes visual observations into a structured latent phase space, evolves this state through Hamiltonian-inspired dynamics, renders future observations with a generative decoder, and uses the resulting rollouts for planning. This architecture is not intended to replace existing world models---whether latent dynamics models, 
which require stable and controllable latent rollouts 
\citep{ha2018worldmodels,hafner2019planet,hafner2020dreamer,hafner2023dreamerv3}, 
or video-generative world models, 
which require physically consistent future synthesis 
\citep{hu2023gaia,bruce2024genie}---but to provide a physical backbone that can improve stability, interpretability, and decision relevance. At a high level, the architecture can be written as
\begin{equation}
    o_{\leq t}
    \xrightarrow[]{E_\theta}
    z_t
    \xrightarrow[]{T_{H}}
    \hat{z}_{t+1:t+H}
    \xrightarrow[]{D_\theta}
    \hat{o}_{t+1:t+H}
    \xrightarrow[]{U}
    a_t^\ast 
\end{equation}, shown in Figure~\ref{fig:hwm_architecture}. Here, $E_\theta$ maps past observations into a latent state, $\mathcal{T}_{H}$ rolls the state forward under future actions, $D_\theta$ renders future observations, and $U$ evaluates the predicted futures for decision making. Such separation is important because modern video generative models can synthesize realistic futures \citep{ho2022videodiffusion,blattmann2023videoldm,hu2023gaia,bruce2024genie}, but physical reliability depends primarily on the structure of the latent dynamics.

The first component is a structured latent representation. Instead of compressing the entire scene into a single global vector, a Hamiltonian World Model represents the scene as a set of entity-level phase variables:
\begin{equation}
    z_t = \{(q_t^i,p_t^i)\}_{i=1}^{N},
\end{equation}
where $q_t^i$ denotes the generalized coordinate of entity $i$, and $p_t^i$ denotes its generalized momentum. The entities may correspond to objects, robot links, tools, surfaces, or dynamically relevant regions. This object-centric representation is motivated by the compositional nature of physical scenes: objects interact through contact, geometry, force, and constraints. Prior work on interaction networks, graph-based simulators, and structured world models suggests that relational representations improve generalization in physical domains \citep{battaglia2016interaction,sanchezgonzalez2020learning,kipf2020contrastive}. In the Hamiltonian view, these entity-level variables provide the substrate on which energy and interaction terms can be defined.

The core transition is governed by a learned Hamiltonian function over the latent phase state:
\begin{equation}
    H_\phi(z_t)
    =
    H_\phi(q_t^1,\ldots,q_t^N,p_t^1,\ldots,p_t^N).
\end{equation}
A natural parameterization decomposes the Hamiltonian into kinetic, potential, and interaction terms:
\begin{equation}
    H_\phi(z_t)
    =
    \sum_i K_\phi(p_t^i)
    +
    \sum_i U_\phi(q_t^i)
    +
    \sum_{i<j} V_\phi(q_t^i,q_t^j).
\end{equation}
This decomposition gives the latent dynamics a physical interpretation: $K_\phi$ captures motion-related energy, $U_\phi$ captures entity-level potential structure, and $V_\phi$ captures interactions between entities. The dynamics are then induced by Hamilton's equations:
\begin{equation}
    \dot{q}_t^i = \frac{\partial H_\phi}{\partial p_t^i},
    \qquad
    \dot{p}_t^i = -\frac{\partial H_\phi}{\partial q_t^i}.
\end{equation}
Equivalently, with $z_t=[q_t,p_t]$,
\begin{equation}
    \dot{z}_t = J\nabla_z H_\phi(z_t),
    \qquad
    J=
    \begin{bmatrix}
        0 & I \\
        -I & 0
    \end{bmatrix}.
\end{equation}
This follows the central insight of Hamiltonian Neural Networks and Hamiltonian Generative Networks: the model learns an energy landscape rather than an arbitrary vector field \citep{greydanus2019hamiltonian,toth2020hamiltonian}. For world modeling, this changes the role of the latent transition from black-box prediction to energy-guided phase-space evolution.

A practical world model, however, cannot assume that real environments are perfectly conservative. Robotics and embodied interaction involve control inputs, friction, damping, contact switches, non-elastic collisions, and perception uncertainty. Therefore, the Hamiltonian transition should be extended to a controlled-dissipative form:
\begin{equation}
    \dot{z}_t
    =
    \left(J - R_\psi(z_t)\right)\nabla_z H_\phi(z_t)
    +
    G_\psi(z_t)a_t
    +
    \epsilon_\psi(z_t,a_t), R_\psi(z_t) \succeq \mathbf{0}.
\end{equation}
The Hamiltonian term provides the physical core, $R_\psi(z_t)$ captures dissipative effects, $G_\psi(z_t)a_t$ models action-conditioned control, and $\epsilon_\psi(z_t,a_t)$ accounts for residual phenomena such as contact discontinuities or unmodeled visual dynamics. This design is consistent with prior work on controlled Hamiltonian learning and symplectic neural dynamics \citep{zhong2020symplectic,hairer2006geometric}. The residual term should remain constrained, for example through
\begin{equation}
    \mathcal{L}_{\mathrm{res}}
    =
    \|\epsilon_\psi(z_t,a_t)\|_2^2,
\end{equation}
so that the model uses the Hamiltonian structure as the dominant explanation of dynamics rather than degenerating into a generic transition network.

The predicted latent trajectory is then passed to a video generative decoder:
\begin{equation}
    p_\theta
    \left(
    o_{t+1:t+H}
    \mid
    o_{\leq t},
    a_{t:t+H-1},
    \hat{z}_{t+1:t+H}
    \right).
\end{equation}

Finally, a Hamiltonian World Model should expose a planning interface. Given a set of candidate action sequences $\{a_{t:t+H-1}^{(m)}\}_{m=1}^{M}$, the model can generate latent and visual rollouts:
\begin{equation}
    \hat{z}_{t+1:t+H}^{(m)}
    =
    T_{H}
    \left(
    z_t,
    a_{t:t+H-1}^{(m)}
    \right),
\end{equation}
\begin{equation}
    \hat{o}_{t+1:t+H}^{(m)}
    \sim
    D_\theta
    \left(
    \hat{z}_{t+1:t+H}^{(m)}
    \right).
\end{equation}
The selected action is then obtained by maximizing a utility function $U$ over predicted futures:
\begin{equation}
m^{*} 
    =
    \arg\max_{m \in \{1,2,3...,M\}} 
    U
    \left(
    \hat{z}_{t+1:t+H}^{(m)},
    \hat{o}_{t+1:t+H}^{(m)}
    \right), a^{*}_{t} = a_{t}^{m^{*}}.
\end{equation}
This is aligned with visual foresight, model predictive control, and latent imagination methods, 
where predicted futures are used to evaluate candidate behaviors before execution 
\citep{camacho2013mpc,finn2017deepvisualforesight,ebert2018visualmpc,hafner2020dreamer,hafner2023dreamerv3}. The difference is that the imagined futures are grounded in a structured phase-space rollout rather than produced by an unconstrained video predictor alone.

The architectural principle is therefore simple: physical evolution and visual rendering should be separated but coupled. The Hamiltonian module provides structured latent dynamics; the video generator provides perceptual detail; the planning module evaluates action usefulness. This makes the model more interpretable than a monolithic video predictor and more suitable for embodied decision making than a purely generative simulator. In this sense, Hamiltonian World Models are not merely video models with a physics-regularized loss; they treat energy-structured latent dynamics as the core mechanism that constrains and guides the generative process.
\vspace{-.2cm}
\section{Distinctions and Challenges of Hamiltonian World Models}
\label{sec:7hwm3}

We treat Hamiltonian World Models as a different design perspective on what a world model ought to represent. From the above Section~\ref{sec:related_work}, existing paradigms can be roughly grouped into four families. \emph{2D video-generative world models} treat the world primarily as a stream of future observations to be synthesized \citep{ho2022videodiffusion,blattmann2023videoldm,hu2023gaia,bruce2024genie}. \emph{3D scene-centric world models} emphasize geometry, view consistency, and explicit spatial structure through radiance fields or scene reconstruction \citep{mildenhall2020nerf,kerbl20233dgs}. \emph{JEPA-like latent world models} focus on predictive abstraction in representation space rather than pixel-level generation, aiming to learn compact and semantically meaningful predictive features \citep{assran2023ijepa,bardes2024vjepa}. By contrast, \emph{Hamiltonian World Models} treat the world as a physically evolving dynamical system in latent phase space, where future observations are rendered from an underlying energy-structured evolution.

Table~\ref{tab:wm_comparison} summarizes the main differences. In brief, 2D video-generative models are strong at perceptual realism and flexible conditional generation, but they often lack explicit physical structure and may drift under long rollouts \citep{hu2023gaia,bruce2024genie}. 3D scene-centric approaches provide stronger spatial grounding and multi-view consistency, but many are better suited to static or quasi-static scene representation than to action-conditioned long-horizon dynamics \citep{mildenhall2020nerf,kerbl20233dgs}. While physical reasoning is not their intended goal, JEPA-like latent models are often weakly tied to explicit physical quantities such as momentum, energy, or contact constraints 
\citep{assran2023ijepa,bardes2024vjepa}. Hamiltonian World Models aim to combine the advantages of structured latent prediction and physical inductive bias by modeling the world in phase space, using energy-based transitions that can, in principle, improve stability, interpretability, and data efficiency \citep{greydanus2019hamiltonian,toth2020hamiltonian,zhong2020symplectic}.
\begin{table*}[t]
\centering
\scriptsize
\caption{Comparison of different world model paradigms.}
\label{tab:wm_comparison}
\renewcommand{\arraystretch}{1.35}
\begin{tabular}{
    >{\centering\arraybackslash}m{3.2cm}
    >{\centering\arraybackslash}m{2.1cm}
    >{\centering\arraybackslash}m{3.0cm}
    >{\centering\arraybackslash}m{3.3cm}
}
\toprule
\textbf{Paradigm} & \textbf{Representation} & \textbf{Strength} & \textbf{Limitation} \\
\midrule

\makecell[c]{
\textbf{2D video-generative}\\
\textbf{world models}\\
\footnotesize e.g., Video Diffusion Models\\
\footnotesize \citep{ho2022videodiffusion}, VideoLDM\\
\footnotesize \citep{blattmann2023videoldm}, GAIA-1\\
\footnotesize \citep{hu2023gaia}, Genie\\
\footnotesize \citep{bruce2024genie}
}
& Image / video sequences
& Strong perceptual realism; flexible conditional future synthesis; intuitive observation-level prediction
& Weak explicit physical structure; prone to long-horizon drift; action consistency may be fragile \\
\midrule

\makecell[c]{
\textbf{3D scene-centric}\\
\textbf{world models}\\
\footnotesize e.g., NeRF\\ \citep{mildenhall2020nerf}\\
\footnotesize 3D Gaussian Splatting\\
\footnotesize \citep{kerbl20233dgs}
}
& Geometry-aware 3D scene representation
& Strong spatial grounding, view consistency, and explicit scene structure
& Reconstruction better than dynamics; limited support for action-conditioned physical evolution \\
\midrule

\makecell[c]{
\textbf{JEPA-like latent}\\
\textbf{world models}\\
\footnotesize e.g., I-JEPA \citep{assran2023ijepa}\\
\footnotesize V-JEPA \citep{bardes2024vjepa}
}
& Predictive latent representation
& Compact and abstract prediction; avoids expensive pixel reconstruction; potentially scalable
& Latent variables are often weakly tied to explicit physics; limited interpretability for control and dynamics \\
\midrule

\makecell[c]{
\textbf{Hamiltonian /}\\
\textbf{physically native}\\
\textbf{world models}
}
& Structured latent representation for phase space $(q,p)$
& Structured physical inductive bias; improved interpretability; potentially better long-horizon stability and data efficiency
& Hard to learn from pixels; real-world dynamics are not fully conservative; contact-rich and deformable phenomena remain difficult \\
\bottomrule
\end{tabular}
\vspace{-.5cm}
\end{table*}
\paragraph{Challenges and boundaries.}
Although Hamiltonian World Models offer an appealing physical perspective, their scope and limitations should be stated clearly. First, real embodied environments are not ideal Hamiltonian systems: they involve friction, control inputs, impacts, dissipation, partial observability, and often non-rigid interactions. A practical model must therefore treat Hamiltonian dynamics as a structural prior rather than a literal description of the full world \citep{zhong2020symplectic,raissi2019physics}. Second, learning phase-space variables from high-dimensional observations remains difficult. Even if a latent state is written as $(q,p)$, inferring meaningful coordinates and momenta from pixels is a nontrivial representation learning problem \citep{toth2020hamiltonian}. Third, contact-rich manipulation, articulated systems, and deformable objects may require hybrid, graph-based, or event-driven extensions beyond smooth Hamiltonian flow \citep{battaglia2016interaction,sanchezgonzalez2020learning}. Finally, current evaluation protocols still provide limited evidence about whether a world model has learned reusable physical regularities. This limitation is also a challenge for Hamiltonian World Models: their value cannot be established merely by lower reconstruction error or higher task reward. Without such evaluations, it remains difficult to distinguish a model that exploits task-specific correlations from one that has learned transferable physical dynamics.

\section{Future Directions}
\vspace{-.2cm}
\label{sec:8}
The broader opportunity is to move from \emph{video world models} toward \emph{physical world foundation models}. In this transition, rather than replacing other paradigms, Hamiltonian World Models are viewed as providing a physically grounded axis along which world models can mature. Future systems may combine the perceptual richness of 2D video generation, the spatial grounding of 3D scene representations, the compact abstraction of JEPA-like latent prediction, and the stability of energy-based phase-space dynamics. A critical future direction is to move from scene-level physical rollout to open-ended physical generalization. Future Hamiltonian World Models may not merely model the interactions among entities in a fixed scene, but should learn reusable physical mechanisms that can be composed across unseen objects, materials, embodiments, and tasks. In this sense, the long-term goal is not only a more stable world model, but a physically native foundation model capable of continual physical abstraction, composition, and self-evolution.

\bibliographystyle{plainnat}
\bibliography{ref1}

\end{document}